\documentclass{article}

\PassOptionsToPackage{square,numbers}{natbib}
\usepackage[final]{neurips_2021}

\setcitestyle{square, numbers}

\usepackage[utf8]{inputenc} 
\usepackage[T1]{fontenc}    
\usepackage{hyperref}       
\usepackage{url}            
\usepackage{booktabs}       
\usepackage{amsfonts}       
\usepackage{nicefrac}       
\usepackage{microtype}      
\usepackage{xcolor}         
\usepackage[pdftex]{graphicx}

\title{Deep learning based landslide density estimation on SAR data for rapid response}

\author{
  Vanessa Boehm \\
  University of California Berkeley \\
  United States \\
  \And
  Wei Ji Leong\\
  The Ohio State University \\
  United States \\
  \And
      Ragini Bal Mahesh \\
  German Aerospace Center DLR \\
  Germany \\
    \And
    Ioannis Prapas \\
  University of Valencia, Spain \\
  National Observatory of Athens, Greece \\
    \And
  Edoardo Nemni \\
  United Nations Satellite Centre\\
  (UNOSAT),Switzerland \\
  \And
  Freddie Kalaitzis \\
  University of Oxford \\
  United Kingdom \\
  \texttt{} \\
  \And
  Siddha Ganju \\
  NVIDIA \\
  United States \\
  \And
  Raul Ramos-Pollan \\
  Universidad de Antioquia \\
  Colombia \\
}

\begin{document}

\maketitle

\begin{abstract}

This work aims to produce landslide density estimates using Synthetic Aperture Radar (SAR) satellite imageries to prioritise emergency resources for rapid response. We use the United States Geological Survey (USGS) Landslide Inventory data annotated by experts after Hurricane María in Puerto Rico on Sept 20, 2017, and their subsequent susceptibility study which uses extensive additional information such as precipitation, soil moisture, geological terrain features, closeness to waterways and roads, etc. Since such data might not be available during other events or regions, we aimed to produce a landslide density map using only elevation and SAR data to be useful to decision-makers in rapid response scenarios.

The USGS Landslide Inventory contains the coordinates of 71,431 landslide heads (not their full extent) and was obtained by manual inspection of aerial and satellite imagery. It is estimated that around 45\% of the landslides are smaller than a Sentinel-1 typical pixel which is 10m $\times$ 10m, although many are long and thin, probably leaving traces across several pixels. Our method obtains 0.814 AUC in predicting the correct density estimation class at the chip level (128$\times$128 pixels, at Sentinel-1 resolution) using only elevation data and up to three SAR acquisitions pre- and post-hurricane, thus enabling rapid assessment after a disaster. The USGS Susceptibility Study reports a 0.87 AUC, but it is measured at the landslide level and uses additional information sources (such as proximity to fluvial channels, roads, precipitation, etc.) which might not regularly be available in an rapid response emergency scenario.

\end{abstract}

\section{Introduction}
According to the United Nations Office for Disaster Risk Reduction, landslides have affected 4.8 million people and caused 18,414 deaths between 1998-2017\footnote{\url{https://www.preventionweb.net/files/61119_credeconomiclosses.pdf}}. Rising temperatures and climate change are projected to worsen this situation~\cite{GarianoLandslidesChangingClimate2016,HuggelPhysicalImpactsClimate2012}, and, given these predictions, there is a growing need for timely and accurate landslide assessment methods that can inform decision-makers and emergency responders.
In the aftermath of a disaster event, optical satellite imagery is commonly used to map the extent of the event. The fact that optical data is often hindered by clouds and limited to daytime observations, motivates the study of weather SAR as a monitoring technique.

The review in \citet{Tehrani_2022} provides a comprehensive view of machine learning and deep learning (DL) methods applied to landslide detection, including the usage of SAR data. Works on DL based landslide density or susceptibility mapping are starting to emerge such as the recent work related to \cite{s22041573} or \cite{Nava_2022}. However, to the best of our knowledge, very few studies are carried out in challenging contexts such as this work, with weak labels, i.e. landslide heads on small landslides.

\paragraph{Application context} This work would enable the rapid generation of coarse landslide density maps after a disaster event using only elevation and SAR data which is commonly available around the globe. These maps can be used to prioritize emergency resources and to validate landslide inventories reducing the need for experts doing fieldwork. Moreover, our approach enables a scenario where landslides are typically small (as compared to SAR acquisition resolution) and the requirement for annotated datasets was relaxed to having weak labels, which are more affordable to produce. 

\section{Data}
\label{sec:data}

\begin{figure}[htbp]
  \centering
  \includegraphics[width=0.7\linewidth]{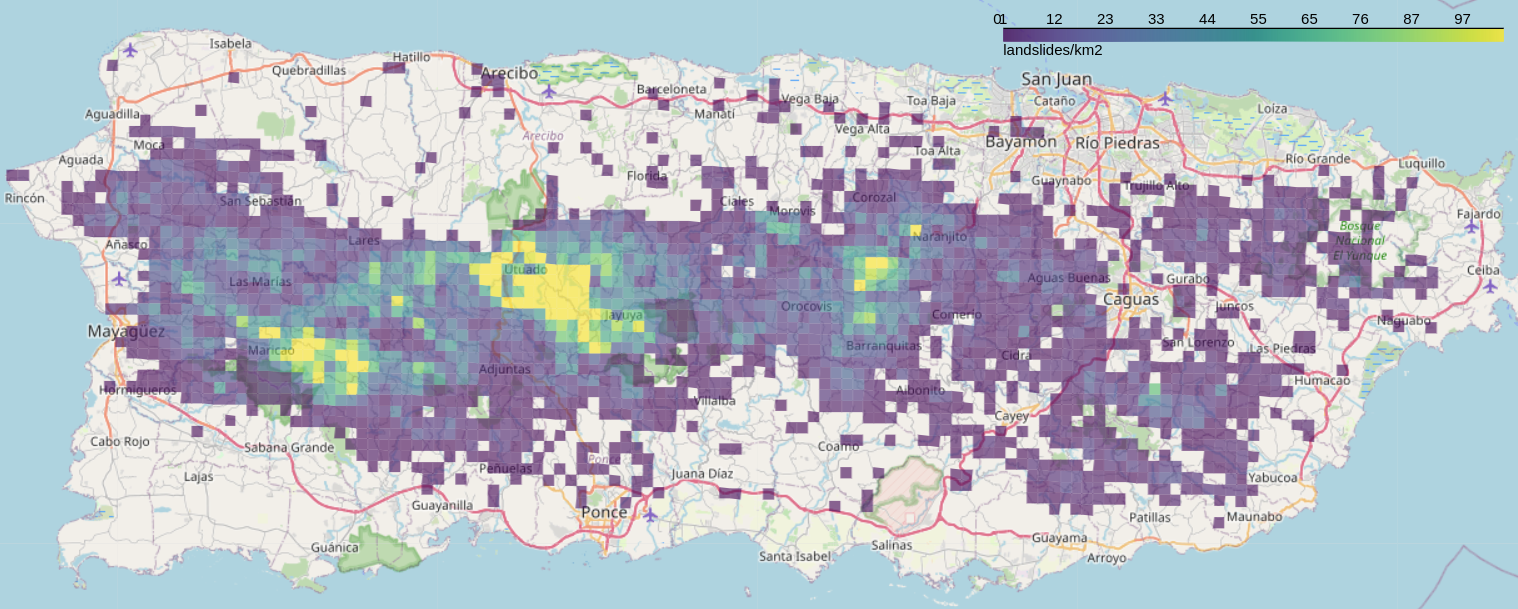}
  \caption{
        Global landslide density, showing 1.28 km$^2$ chips with 1 or more landslides
  }
  \label{fig:globaldensity}
\end{figure}

\paragraph{Landslide labels}
We used the landslide inventory dataset produced by the USGS on slope failure locations in Puerto Rico after Hurricane María \cite{https://doi.org/10.5066/p9bvmd74}, which is also used to validate their landslide susceptibility map on the island \cite{Hughes2020}. The dataset contains 71,431 landslides annotated by a group of experts who marked the head of each landslide by inspecting optical imagery from different sources \cite{Hughes2020}. Figure \ref{fig:globaldensity} shows the overall density of landslides obtained from this dataset.

\begin{figure}[htbp]
  \centering
  \includegraphics[width=0.9\linewidth]{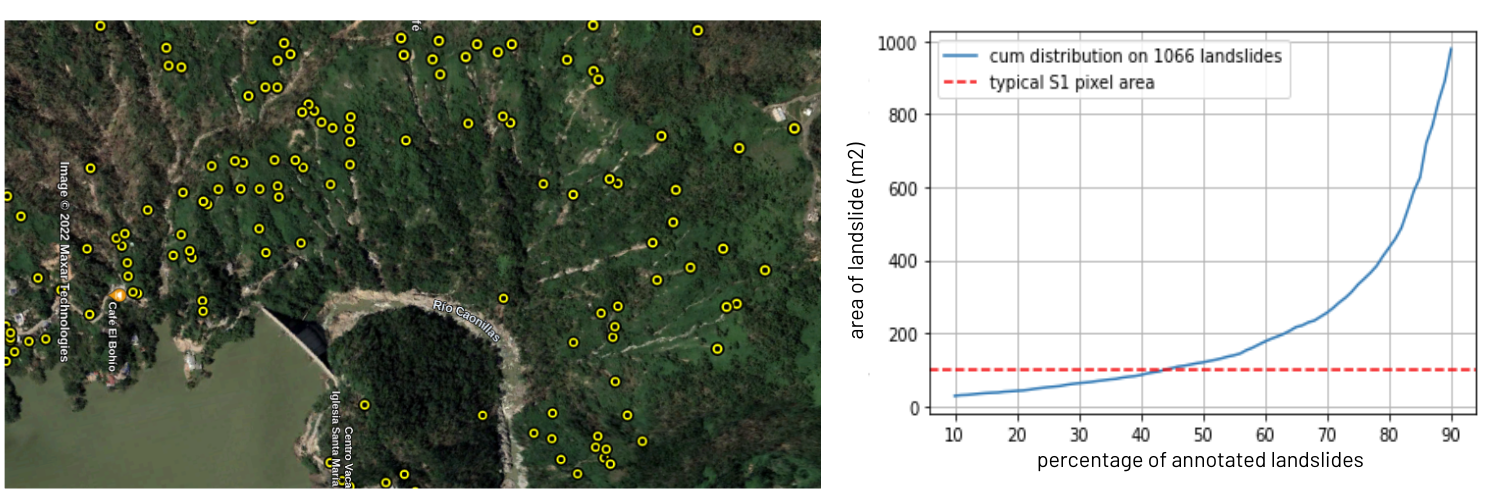}
  \caption{
        Left: example landslide heads annotations around Lago Caonillas in the Utuado region. Right: size distribution of 1'066 landslides.
  }
  \label{fig:landslidedistribution}
\end{figure}

Additionally, the USGS produced a dataset of four selected regions in Puerto Rico annotation the landslides contours \cite{https://doi.org/10.5066/p9ow4slx} \cite{bessette2019landslides}. This accounts for 1,066 landslides, which we used only to get a better sense of the nature and extent of the landslides. As shown in Figure \ref{fig:landslidedistribution}, most landslides occupy an area smaller than a Sentinel-1 typical pixel, but are thin and long, probably leaving traces across several pixels. This makes this problem specially challenging.

\paragraph{Input data}
We set forth to produce a discretized density estimation by using SAR polarimetry and a digital elevation model (DEM).  As detailed in table \ref{table:s1data}, we used SAR acquisitions from Sentinel-1 pre and post Hurricane Maria as downloaded from Microsoft Planetary Computing service\footnote{https://planetarycomputer.microsoft.com/}. For elevation data we used the SRTM 1 Arc-second global \cite{https://doi.org/10.5066/f7pr7tft} DEM and used the \texttt{richdem} Python package \cite{RichDEM} to generate derived elevation measures (slope, curvature and aspect).

\begin{table}
  \caption{Sentinel-1 acquistions used. Hurricane Maria hit Puerto Rico on Sep 20, 2017, so we have three acquisitions pre-event and three post-event. All acquisitions are ascending which ensures comparable SAR geometry.}
  \label{table:s1data}
  \centering
  \begin{tabular}{lllllll}
    \toprule
    date     & satellite     & id & orbit & path & frame & direction \\
    \midrule
20170823T223637 (pre)	& S1A	& 9D3C	& 018057	& 135	& 55	& ascending \\
20170904T223637	(pre)& S1A	& B225	& 018232	& 135	& 55	& ascending \\
20170916T223638	(pre)& S1A	& AE4E	& 018407	& 135	& 55	& ascending \\
20170922T223544	(pos)& S1B	& D309	& 007511	& 135	& 52	& ascending \\
20171010T223638 (pos)& S1A	& 99FA	& 018757	& 135	& 55	& ascending \\
20171022T223638	(pos)& S1A	& 26BB	& 018932	& 135	& 55	& ascending \\ 
    
    \bottomrule
  \end{tabular}
\end{table}

\begin{figure}[htbp]
  \centering
  \includegraphics[width=1.0\linewidth]{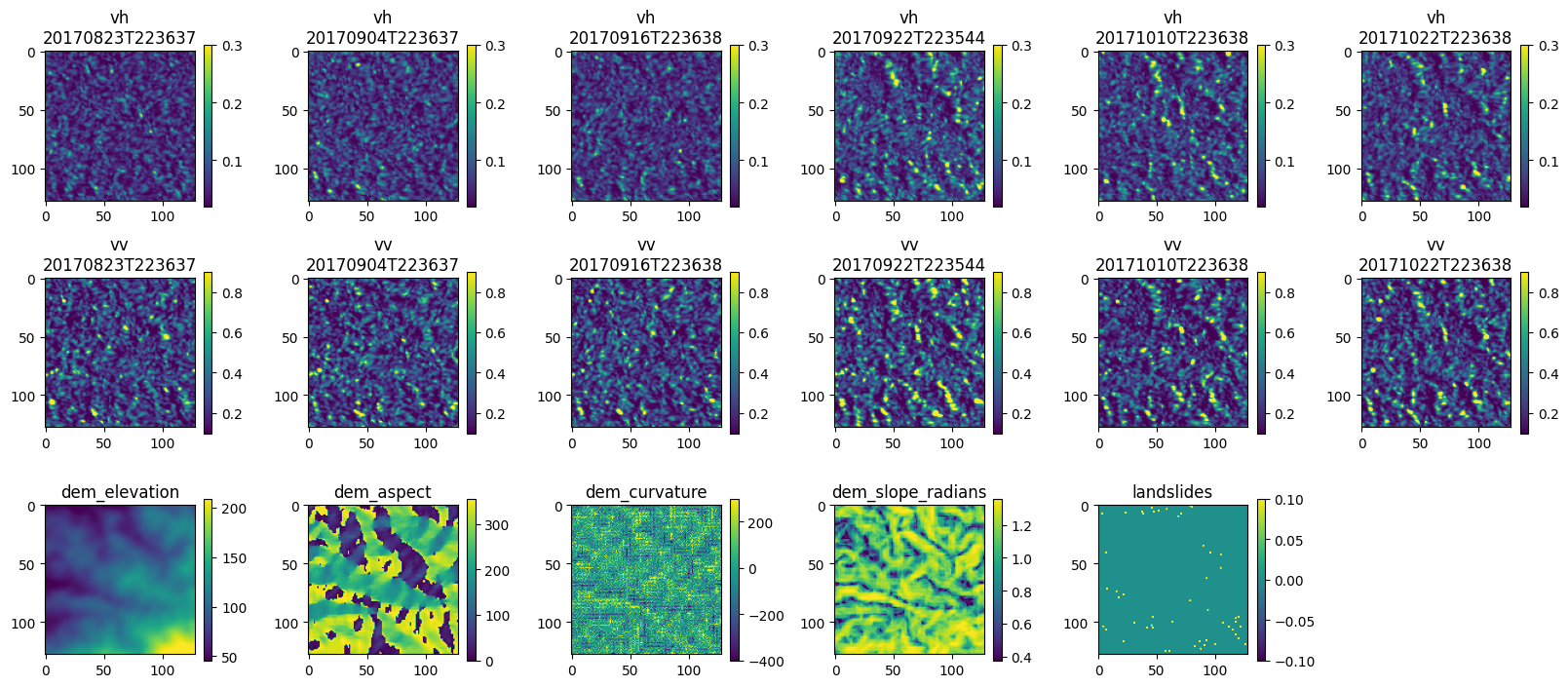}
  \caption{
        Example of one chip. Note how the  backscattering intensity increased after Hurricane María on Sept 20, 2017.
  }
  \label{fig:chips}
\end{figure}

\paragraph{Chips generation}
Chips are \texttt{numpy} arrays of size 128$\times$128 with 17 channels corresponding to 4 DEM channels (elevation, slope, curvature, aspect), 2 SAR channels (VV + VH) for each of the 6 Sentinel-1 revisits, and 1 channel for the labels. Pixels in the label channel have a value of 1 when there is a landslide head annotated in that location, and zero otherwise. In total, we produced 5,517 chips of dimensions 128$\times$128 with a spatial resolution of 10m$\times$10m covering 9,039 km$^2$ which corresponds to the surface area of Puerto Rico. Figure \ref{fig:chips} shows an example chip.


\section{Methodology}
\label{sec:methodology}

\paragraph{Model}
\label{subsec:model}

Following the methodology in \citet{Hughes2020} our end goal is to classify each chip in one of the following \emph{density classes}: 0 for a chip with no landslides, 1 for a chip containing between 1 and 25 landslides/km$^2$ and 2 if the chip contains more than 25 landslides/km$^2$.

\begin{figure}[htbp]
  \centering
  \includegraphics[width=1.0\linewidth]{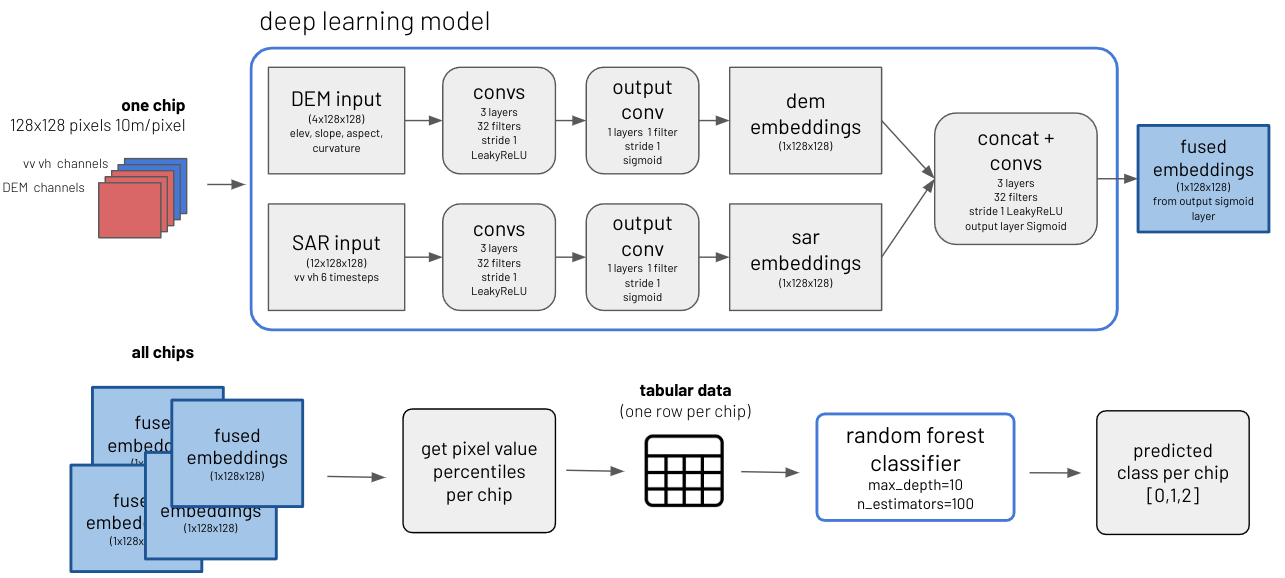}
  \caption{
        Deep Learning model architecture and classification pipeline with Random Forest
  }
  \label{fig:models}
\end{figure}

We propose a two-stage approach as illustrated in Figure \ref{fig:models}. We trained a convolutional DL model on SAR+DEM input to produce, for each chip, a mask of the same dimensions (128x128) with pixel values $\in [0,1]$. This mask is the \emph{fused embeddings} that the model will produce for a chip. The intent is that embeddings on pixels with landslides will have a higher value than other pixels. However, as opposed to a segmentation task, we devise a loss function, described below, which we believe more adequate to our end task of density estimation at a chip level. Second, for each chip embedding, we compute a vector with 19 elements containing the percentiles 5\%, 10\%, ... 95\% of pixel embeddings values. Thus, after inference through the DL model, each chip gets described with a vector. Together with the associated class of each chip, we feed this to a random forest classifier to get the final density class prediction.

 All activation maps from end to end are 128x128 in dimensions. Also, the output layer is defined with a sigmoid activation so that the embeddings values are $\in [0,1]$. Finally, notice that there are two independent convolutional branches to deal with DEM and SAR channels as if they were different modalities. Preliminary experiments showed that this architecture is preferred for dealing with all channels together.

\paragraph{Loss function}
For the DL model, and in order to deal with the challenges of the landslide head labels, we defined the following loss function at the batch level. Recall that for a batch of $n$ chips as input the model will produce $n$ embeddings of 128x128 pixels each. We denote with $e^+_i$ the set of values of pixels of embedding $i$ that fall on pixels marked with a landslide head, and with $e^-_i$ we denote the rest. We gather all pixel values in the batch on landslides under $E^+ = \bigcup\limits_{i=1}^{n}e^+_{i}$ and the rest under $E^- = \bigcup\limits_{i=1}^{n}e^-_{i}$. Then, our loss for a batch is

\begin{equation}
\label{eq:loss}
\mathcal{L}_{batch} =  (1 - \overline{E^+})  + \overline{E^-}
\end{equation}

where $\overline{E^+}$ and $\overline{E^-}$ denote the average value of all elements of $E^+$ and $E^-$, which value will always be $\in [0,1]$ as per a sigmoid output. Intuitively,  we want embedding pixels on landslide labels to have in \emph{average} a high value, and the opposite for the rest of the pixels. Observe that, usually, there are many more pixels without landslides so $E^-$ will contain many more elements than $E^+$. For instance, in a batch of 8 chips with 100 landslides in total, $E^+$ will contain 100 values and $E^-$ will contain $128\times 128\times 8-100=130,972$ values

Since only the landslide heads are annotated, many landslide pixels will be included in $E^-$, and not in $E^+$. However, since $E^-$ is much more numerous, their value will get diluted and will not penalize the loss. This loss will tend to generate many false positives at the pixel level, but in a way that we expect to benefit the final density estimation task. See Figure \ref{fig:results} to get a sense of this.

\paragraph{Experimental setup}
\label{subsec:experimental_setup}
Since we want to produce a prediction map for the full island of Puerto Rico, we split all chips randomly into two sets of equal size. Then, we run the training and prediction pipeline twice. First we train and run the model selection with split 1 (40\% train, 10\% validation) and generate predictions over split 2 (50\% of the data). Then, we trained on split 2 (also subdivided in 40/10) and generated predictions on split 1.

We devised a set of configurations to understand our pipeline behaviour with respect to two aspects. First, we run configurations using only DEM input, only SAR input and both fused, using only the relevant part of the architecture in Figure \ref{fig:models}. Second, we wanted to measure the effect of additional revisits after hurricane Maria in the model results so we run models using data for the first, second and third revisits after the event corresponding to the last three lines (after Sep 20, 2017) in Table \ref{table:s1data}.

\section{Results}
\label{sec:results}
\begin{figure}[htbp]
  \centering
  \includegraphics[width=1.0\linewidth]{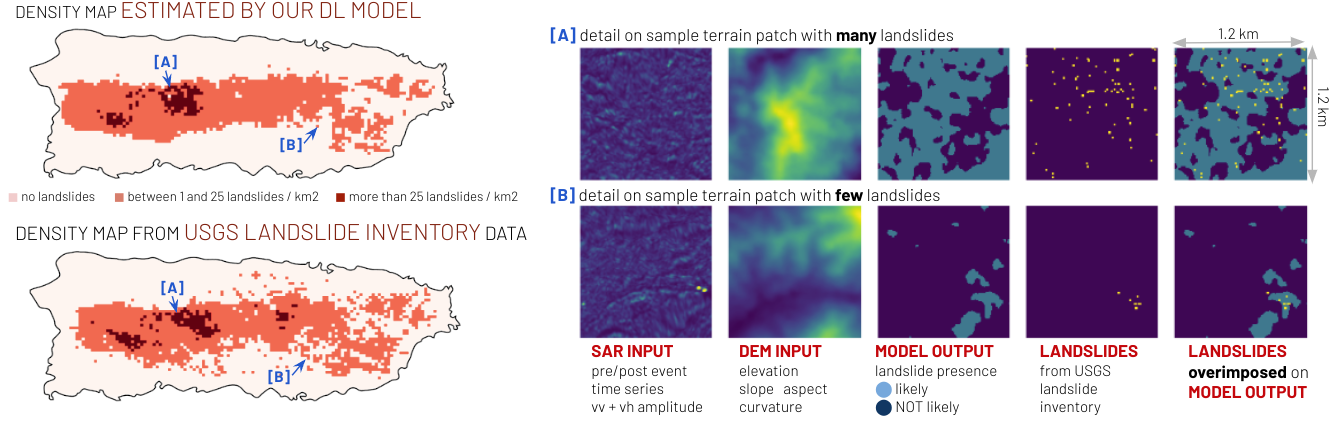}
  \caption{
        Resulting density map and example embeddings. 
  }
  \label{fig:results}
\end{figure}

Figure \ref{fig:results} shows the obtained three-class density map with three revisits using DEM+SAR input data, together with the embeddings and landslides for two chips. We observed that landslides fall mostly on light blue embeddings which was the intention. Pixel-wise, there is a large false positive rate, many light blue pixels with no landslides heads, but this is expected as the loss function \ref{eq:loss} dilutes the penalty on these cases by design. This is aligned with the fact that density estimation by the random forest is roughly done by taking the average embedding value for each chip, a larger portion of light blue coverage indicates a larger landslide density on that chip. 

Table \ref{table:auc} details the results for each experiment configuration. Our method gets 0.782 AUC using the first revisit two days after the hurricane, and up to 0.814 when using three satellite revisits, one month after the event. It is interesting to see that DEM-only models get an AUC of around 0.7, signalling the correlation of the landslides with terrain features. Moreover, fusing SAR and DEM inputs definitely boosts the model performance. We compare our results with the USGS aggregated susceptibility index modified by soil moisture (SIA$_m$) \cite{Hughes2020} which, as mentioned, uses a weighted mixture of a variety of input data and is measured on the same landslide inventory dataset.

\begin{table}
  \caption{Results summary. USGS SIA$_m$ is the aggregated susceptibility index}
  \label{table:auc}
  \centering
  \begin{tabular}{llllllll}
    \toprule
     pos event data
     & \multicolumn{2}{c}{1 revisit}     
     & \multicolumn{2}{c}{2 revisits} 
     & \multicolumn{2}{c}{3 revisits} & \\

    \cmidrule(r){2-3}
    \cmidrule(r){4-5}
    \cmidrule(r){6-7}

    experiment & train     & val     & train & val & train & val &  \\
    \midrule
    DEM only     & 0.705   & 0.698 &  0.699  & 0.687  &  0.701  & 0.692  &  \\
    SAR only     & 0.710   & 0.708      &  0.715 & 0.710  &  0.742 & 0.740  &  \\
    DEM + SAR fused     & 0.801   & \textbf{0.782}     & 0.828  & \textbf{0.805}  & 0.831  & \textbf{0.814}  &  \\
    \midrule
    USGS SIA$_m$     &    &      &   &   &   &   &  0.87\\
    \bottomrule
  \end{tabular}
\end{table}

\section{Conclusions}
\label{sec:conclusions}
This work shows that coarse density estimation using weak labels on small landslides is feasible in a rapid response context, using only elevation and SAR data. From here on, we believe there are several avenues towards having an operational SAR based density estimation service. These include better understanding the model's response to data (Figure \ref{fig:chips} seems to show that SAR is responding to many things after the hurricane not only landslides), measuring the transferability of models across different regions, validating the interpretability for experts of the embeddings produced, and more.

\section{Acknowledgements}
This work was funded by the FDL/NASA 2022 research sprint. For their invaluable insights we are indebt to Aaron Piña and Gerald Bawden (NASA SMD), Eric Fielding, Erica Podest and Alex Handwerger (NASA JPL), John Stock, Jason Stocker and Corina Cerovski-Darriau (USGS), Forrest Williams and Franz Mayer (Alaska Satellite Facility), Ioannis Papoutsis (National Greek Observatory), Ronny Haensch (DLR) and Brad Neuberg (Planet).

\bibliography{references}

\begin{thebibliography}{11}
\providecommand{\natexlab}[1]{#1}
\providecommand{\url}[1]{\texttt{#1}}
\expandafter\ifx\csname urlstyle\endcsname\relax
  \providecommand{\doi}[1]{doi: #1}\else
  \providecommand{\doi}{doi: \begingroup \urlstyle{rm}\Url}\fi

\bibitem[Gariano and Guzzetti(2016)]{GarianoLandslidesChangingClimate2016}
Stefano~Luigi Gariano and Fausto Guzzetti.
\newblock Landslides in a changing climate.
\newblock \emph{Earth-Science Reviews}, 162:\penalty0 227--252, November 2016.
\newblock ISSN 00128252.
\newblock \doi{10.1016/j.earscirev.2016.08.011}.
\newblock URL
  \url{https://linkinghub.elsevier.com/retrieve/pii/S0012825216302458}.

\bibitem[Huggel et~al.(2012)Huggel, Khabarov, Korup, and
  Obersteiner]{HuggelPhysicalImpactsClimate2012}
Christian Huggel, Nikolay Khabarov, Oliver Korup, and Michael Obersteiner.
\newblock Physical impacts of climate change on landslide occurrence and
  related adaptation.
\newblock In John~J. Clague and Douglas Stead, editors, \emph{Landslides},
  pages 121--133. Cambridge University Press, 1 edition, August 2012.
\newblock ISBN 978-0-511-74036-7 978-1-107-00206-7.
\newblock \doi{10.1017/CBO9780511740367.012}.
\newblock URL
  \url{https://www.cambridge.org/core/product/identifier/9780511740367%23c00206-11-1/type/book_part}.

\bibitem[Tehrani et~al.(2022)Tehrani, Calvello, Liu, Zhang, and
  Lacasse]{Tehrani_2022}
Faraz~S. Tehrani, Michele Calvello, Zhongqiang Liu, Limin Zhang, and Suzanne
  Lacasse.
\newblock Machine learning and landslide studies: recent advances and
  applications.
\newblock \emph{Nat Hazards}, jun 2022.
\newblock \doi{10.1007/s11069-022-05423-7}.
\newblock URL \url{https://doi.org/10.1007%2Fs11069-022-05423-7}.

\bibitem[Ghasemian et~al.(2022)Ghasemian, Shahabi, Shirzadi, Al-Ansari,
  Jaafari, Kress, Geertsema, Renoud, and Ahmad]{s22041573}
Bahareh Ghasemian, Himan Shahabi, Ataollah Shirzadi, Nadhir Al-Ansari, Abolfazl
  Jaafari, Victoria~R. Kress, Marten Geertsema, Somayeh Renoud, and Anuar
  Ahmad.
\newblock A robust deep-learning model for landslide susceptibility mapping: A
  case study of kurdistan province, iran.
\newblock \emph{Sensors}, 22\penalty0 (4), 2022.
\newblock ISSN 1424-8220.
\newblock \doi{10.3390/s22041573}.
\newblock URL \url{https://www.mdpi.com/1424-8220/22/4/1573}.

\bibitem[Nava et~al.(2022)Nava, Bhuyan, Meena, Monserrat, and
  Catani]{Nava_2022}
Lorenzo Nava, Kushanav Bhuyan, Sansar~Raj Meena, Oriol Monserrat, and Filippo
  Catani.
\newblock Rapid mapping of landslides on {SAR} data by attention u-net.
\newblock \emph{Remote Sensing}, 14\penalty0 (6):\penalty0 1449, mar 2022.
\newblock \doi{10.3390/rs14061449}.
\newblock URL \url{https://doi.org/10.3390%2Frs14061449}.

\bibitem[Hughes et~al.(2019)Hughes, Bayouth~Garcia, Martinez~Milian, Schulz,
  and Baum]{https://doi.org/10.5066/p9bvmd74}
Kenneth~Stephen Hughes, Desiree Bayouth~Garcia, Gabriel Martinez~Milian,
  William~H Schulz, and Rex~L Baum.
\newblock Map of slope-failure locations in puerto rico after hurricane maria,
  2019.
\newblock URL
  \url{https://www.sciencebase.gov/catalog/item/5d4c8b26e4b01d82ce8dfeb0}.

\bibitem[Hughes and Schulz(2020)]{Hughes2020}
K.~Stephen Hughes and William Schulz.
\newblock Map depicting susceptibility to landslides triggered by intense
  rainfall, puerto rico, 2020.
\newblock URL \url{https://doi.org/10.3133/ofr20201022}.

\bibitem[Bessette-Kirton et~al.(2019{\natexlab{a}})Bessette-Kirton, Coe,
  Cerovski-Darriau, Kelly, and Schulz]{https://doi.org/10.5066/p9ow4slx}
Erin~K. Bessette-Kirton, Jeffrey~A Coe, Corina Cerovski-Darriau, Matthew~A.
  Kelly, and William~H. Schulz.
\newblock Map data from landslides triggered by hurricane maria in four study
  areas of puerto rico, 2019{\natexlab{a}}.
\newblock URL
  \url{https://www.sciencebase.gov/catalog/item/5ca3c65fe4b0b8a7f6334309}.

\bibitem[Bessette-Kirton et~al.(2019{\natexlab{b}})Bessette-Kirton,
  Cerovski-Darriau, Schulz, Coe, Kean, Godt, Thomas, and
  Hughes]{bessette2019landslides}
Erin~K Bessette-Kirton, Corina Cerovski-Darriau, William~H Schulz, Jeffrey~A
  Coe, Jason~W Kean, Jonathan~W Godt, Matthew~A Thomas, and K~Stephen Hughes.
\newblock Landslides triggered by hurricane maria: Assessment of an extreme
  event in puerto rico.
\newblock \emph{GSA Today}, 29\penalty0 (6):\penalty0 4--10,
  2019{\natexlab{b}}.

\bibitem[{Earth Resources Observation And Science (EROS)
  Center}(2017)]{https://doi.org/10.5066/f7pr7tft}
{Earth Resources Observation And Science (EROS) Center}.
\newblock Shuttle radar topography mission (srtm) 1 arc-second global, 2017.
\newblock URL
  \url{https://www.usgs.gov/centers/eros/science/usgs-eros-archive-digital-elevation-shuttle-radar-topography-mission-srtm-1-arc?qt-science_center_objects=0#qt-science_center_objects}.

\bibitem[Barnes(2016)]{RichDEM}
Richard Barnes.
\newblock \emph{RichDEM: Terrain Analysis Software}, 2016.
\newblock URL \url{http://github.com/r-barnes/richdem}.

\end{thebibliography}
\newpage

\end{document}